\documentclass{article}
\usepackage{ijcai25}

\usepackage{times}
\usepackage{textcomp}

\usepackage{soul}
\usepackage{url}
\usepackage[hidelinks]{hyperref}
\usepackage[utf8]{inputenc}
\usepackage[small]{caption}
\usepackage{graphicx}
\usepackage{amsmath}
\usepackage{amsthm}
\usepackage{amsfonts}
\usepackage{booktabs}
\usepackage{algorithm}
\usepackage{algorithmic}
\usepackage[switch]{lineno}
\usepackage{multirow}

\usepackage[table]{xcolor}  

\title{Skin-SOAP: A Weakly Supervised Framework for Generating Structured SOAP Notes}

\author{
Sadia Kamal$^1$ \and
Tim Oates$^1$ \and
Joy Wan$^2$ \\
\affiliations
$^1$Department of Computer Science, University of Maryland, Baltimore County \\
$^2$Department of Dermatology, Johns Hopkins University School of Medicine, Baltimore, MD \\
\emails
sadia1402@umbc.edu, oates@cs.umbc.edu, jwan7@jhmi.edu
}

\begin{document}

\maketitle

\begin{abstract}
Skin carcinoma is the most prevalent form of cancer globally, accounting for over \$8 billion in annual healthcare expenditures. Early diagnosis, accurate and timely treatment are critical to improving patient survival rates. In clinical settings, physicians document patient visits using detailed SOAP (Subjective, Objective, Assessment, and Plan) notes. However, manually generating these notes is labor-intensive and contributes to clinician burnout. In this work, we propose \textit{\textbf{skin-SOAP}}, a weakly supervised multimodal framework to generate clinically structured SOAP notes from limited inputs, including lesion images and sparse clinical text. Our approach reduces reliance on manual annotations, enabling scalable, clinically grounded documentation while alleviating clinician burden and reducing the need for large annotated data. Our method achieves performance comparable to GPT-4o, Claude, and DeepSeek Janus Pro across key clinical relevance metrics. To evaluate this clinical relevance, we introduce two novel metrics MedConceptEval and Clinical Coherence Score (CCS) which assess semantic alignment with expert medical concepts and input features, respectively.



\end{abstract}

\section{Introduction}

Skin cancer remains one of the most common and deadliest cancers in the United States, with approximately 9,500 new cases diagnosed daily \cite{rogers2015incidence}, highlighting the crucial role of clinical documentation as the foundation for effective communication, accurate diagnosis, and informed treatment planning. Structured formats like SOAP (Subjective, Objective, Assessment, Plan) notes are widely adopted in the United States to ensure consistency in recording patient encounters and minimizing communication errors among healthcare professionals \cite{schloss2020towards}. However, generating these notes is labor-intensive and time-consuming, which reduces direct patient interaction time and significantly contributes to physician burnout \cite{li2024improving} \cite{biswas2024intelligent}.

Automating the generation of SOAP notes presents a promising solution to reduce administrative burden, improve documentation consistency, and allow clinicians to focus more on patient centred care. Recent advances in large language models (LLMs) have enabled impressive progress in medical natural language processing tasks, including clinical summarization, question answering \cite{singhal2023large}, lab report interpretation \cite{he2024pathclip}, and deidentification of sensitive information \cite{yang2023large}. These models can produce coherent and fluent clinical narratives, making them useful tools for medical documentation. However, general purpose LLMs often lack the domain-specific reasoning required for clinical settings, struggle to understand subtle medical context, and are generally limited to text based inputs. Their performance is further constrained in tasks such as structured note generation, especially when applied to domains like dermatology.


\begin{figure*}[t]
    \centering
    \includegraphics[width=0.8\textwidth]{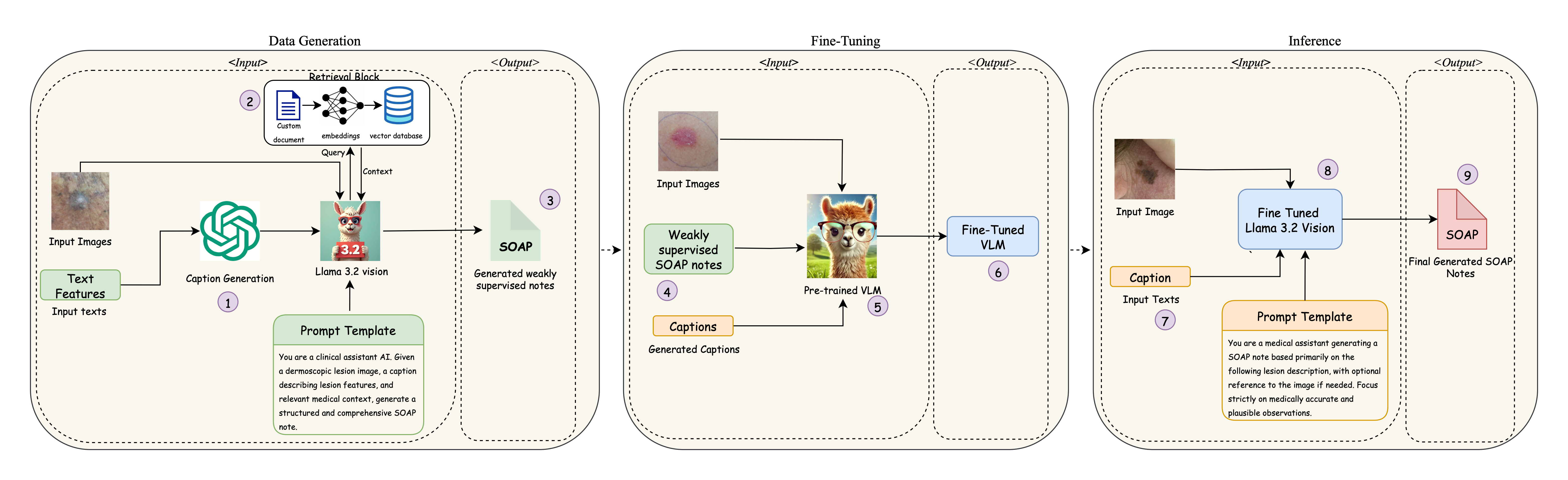}
    \caption{Overview of the proposed skin-SOAP framework, consisting of data generation, fine-tuning, and inference phases.}
    \label{fig:onecol}
\end{figure*}

Existing methods for automated SOAP note generation, such as K-SOAP \cite{li2024improving}, rely heavily on extensive doctor-patient dialogues and large-scale annotated datasets resources that are particularly limited in dermatology and skin lesion documentation \cite{wei2024artificial}. Moreover, capturing both the visual features of skin conditions and the underlying clinical reasoning in a structured format remains a major challenge. To address these limitations, we propose skin-SOAP, a novel weakly supervised multimodal framework that generates structured SOAP notes from limited inputs including lesion images and sparse clinical text. Unlike prior approaches focused solely on text-based SOAP generation or dermatologic diagnosis, our method uniquely integrates retrieval-augmented clinical knowledge, weak supervision, and multimodal synthesis to enable domain-aligned documentation without requiring large-scale annotations. We also introduce two novel metrics to evaluate clinical and semantic quality: MedConceptEval and Clinical Coherence Score (CCS), which go beyond traditional NLP metrics by assessing alignment with defined concepts and feature consistency. By leveraging domain-guided retrieval and pseudo-labeling, our framework produces clinically relevant SOAP notes with minimal supervision, offering a scalable solution for dermatology and broader healthcare applications.
\textbf{Our main contributions are:}
\begin{itemize}
    \item We developed skin-SOAP, a weakly supervised multimodal framework for structured SOAP note generation from lesion images and limited clinical text.
    \item We introduced two novel evaluation metrics: MedConceptEval and Clinical Coherence Score (CCS) to evaluate semantic alignment with clinical concepts and input consistency.
    \item We performed statistical analysis using two-way ANOVA to quantify the effects of SOAP sections and lesion types on semantic similarity scores.
    \item We conducted qualitative evaluation using an LLM-as-a-Judge framework (Flow-Judge-v0.1) to assess structure, readability, completeness, and clinical relevance.
\end{itemize}

\section{Related Work}
\label{sec:related}

Artificial intelligence has been widely applied in the medical domain for tasks such as medical image captioning \cite{cheng2025caparena}, clinical text summarization \cite{van2023clinical}, discharge note generation \cite{jung2024enhancing}, and medical question answering \cite{yan2024large}. Recent efforts have focused on generating clinical notes, especially SOAP notes, from Electronic Health Records (EHRs) and doctor-patient conversations. In \cite{li2024improving}, large language models (LLMs) are fine-tuned to generate K-SOAP notes, while \cite{chen2024exploring} evaluated generative models across general and SOAP-specific formats. Similar work has explored domains like pediatric rehabilitation \cite{amenyo2025assessment} and fine-tuning LLMs for efficient SOAP note generation \cite{leong2024efficient}.

Earlier approaches to SOAP note generation, such as \cite{schloss2020towards,singh2023large}, used sequence-to-sequence models and transformer architectures trained on large dialogue datasets. Although general-purpose LLMs have improved clinical NLP tasks \cite{singhal2023large,yang2023large}, they often lack domain-specific reasoning and structured multimodal output capabilities.  While methods like \cite{ramprasad2023generating} have attempted to improve faithfulness in text-based settings, these methods heavily rely on large annotated dialogue corpora, which are difficult to obtain due to privacy concerns \cite{chen2024exploring}, and they remain text-centric without leveraging multimodal information, which is critical to fields like dermatology. SkinGPT-4 \cite{zhou2024pre} recently demonstrated that integrating clinical images and text using a multimodal LLM improves dermatological diagnostic reasoning. However, it is primarily designed for diagnostic prediction and lacks structured documentation capabilities. Specifically, it does not generate clinically formatted SOAP notes or support section-level reasoning required for clinical documentation. In contrast, our work focuses on end-to-end generation of structured SOAP notes from limited multimodal inputs, introducing weak supervision and domain-guided retrieval to ensure both clinical reliability and scalability.

\section{Methodology}
\label{sec:approach}

We proposed a three-phase weakly supervised multimodal framework for generating clinically structured SOAP notes from limited dermatologic inputs. As illustrated in Fig.~\ref{fig:onecol}, our method consists of : (1) a data generation phase to synthesize weakly supervised SOAP notes using generative captioning and retrieval-augmented knowledge integration, (2) a fine-tuning phase to adapt a vision-language model using the synthesized notes, and (3) an inference phase to generate high-quality structured SOAP notes from new patient data.

\subsection{Dataset}
\label{subsec:data}

We use the PAD-UFES-20 dataset~\cite{pacheco2020pad}, which consists of 2,298 dermoscopic images along with structured metadata for 1,641 skin lesions collected from 1,373 patients. The lesions are classified into six types: Basal Cell Carcinoma (BCC), Melanoma (MEL), Squamous Cell Carcinoma (SCC, including Bowen’s disease), and three non-cancerous conditions: Actinic Keratosis (ACK), Seborrheic Keratosis (SEK), and Nevus (NEV).

Approximately 58\% of the samples are confirmed through biopsy, while the remaining cases are clinically diagnosed based on expert consensus. Each lesion is linked to a CSV file containing 26 structured clinical features, which include lesion characteristics (such as size and anatomical location), patient demographics, symptom information (such as itching, bleeding, or changes in appearance), and family medical history.

\subsection{Data Generation}
\label{subsec:pretraining}

Due to the limited availability of large-scale annotated SOAP note datasets in dermatology, we employ a weak supervision strategy to synthesize training data. Each training sample consists of a lesion image paired with structured clinical features, such as lesion diameter, biopsy status, and symptom descriptors. We first use GPT-3.5~\cite{brown2020language} to generate a clinical caption summarizing these structured attributes into a coherent description of the lesion. To improve the clinical relevance and factual reliability of the generated notes, we design a retrieval-augmented generation framework \cite{lewis2020retrieval}. The generated caption is used as a query to retrieve semantically relevant passages from a curated vector database built with ChromaDB. This database indexes document chunks extracted from authoritative medical sources, including the South Texas Skin Cancer Institute \cite{stxskincancer}, the National Cancer Institute \cite{cancer_gov}, the American Cancer Society \cite{american_cancer_society}, and the UK's National Health Service \cite{nhs_melanoma_symptoms}, covering lesion types, diagnostic criteria, symptomatology, and treatment guidelines. The retrieved context are concatenated with the original caption and provided as input to the pre-trained Vision-LLaMA 3.2 model, guided by a structured prompting template that encourages SOAP format outputs. This design addresses common limitations of pre-trained language models, such as outdated knowledge and hallucinated reasoning, and enables the model to generate reliable and clinically grounded, weakly supervised SOAP notes.

\subsection{Fine-Tuning}
\label{subsec:finetuning}

\begin{figure}[htbp]
    \centering
    \includegraphics[width=0.4\textwidth]{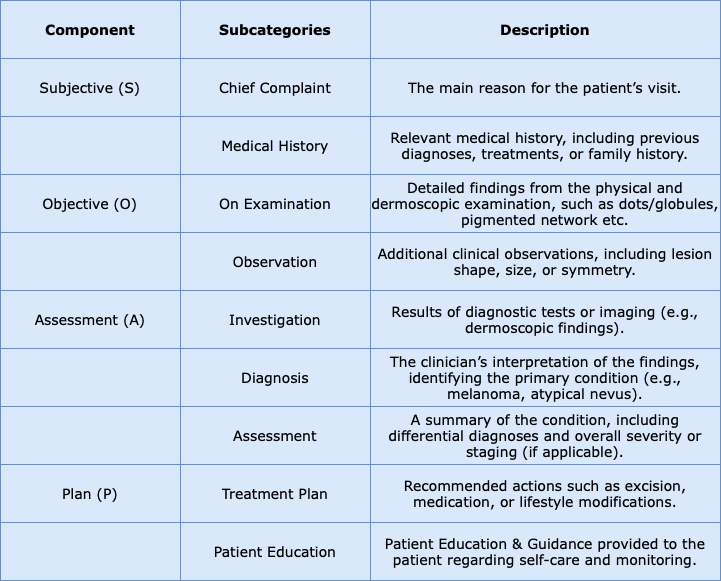}
    \caption{Structured SOAP note components.}
    \label{fig:SOAP_table}
\end{figure}

We fine-tune the Vision-LLaMA 3.2 model using the synthesized dataset, where the lesion image and generated caption are treated as multimodal inputs, and the weakly supervised SOAP note serves as the training target. The model is optimized to produce structured outputs following the standard SOAP note format illustrated in Fig ~\ref{fig:SOAP_table}.

\subsubsection{Parameter-Efficient Fine-Tuning (PEFT)}
\label{subsubsec:peft}

To reduce computational costs without compromising model performance, we employ Parameter-Efficient Fine-Tuning (PEFT) strategies. Specifically, we use Quantized Low-Rank Adaptation (QLoRA) ~\cite{dettmers2023qlora}, which introduces trainable low-rank matrices into transformer layers. Instead of updating all model parameters, QLoRA injects low-rank decompositions into specific modules such as the query, key, value, and output projections. The low-rank adaptation update is defined as:





\begin{equation}
\Delta W = AB
\label{eq:deltaW}
\end{equation}

\noindent where \(A \in \mathbb{R}^{d \times r}\) and \(B \in \mathbb{R}^{r \times d}\) are the learnable low-rank matrices with \(r \ll d\).

The adapted weight matrix during fine-tuning becomes:

\begin{equation}
W' = W + \Delta W = W + AB
\label{eq:adaptedW}
\end{equation}

\noindent
where \(W\) represents the original frozen weights, and \(\Delta W\) captures the trainable adaptation.
We apply LoRA to the \texttt{q}, \texttt{k}, \texttt{v}, \texttt{o}, \texttt{gate}, \texttt{up}, and \texttt{down} modules of the Vision-LLaMA model, allowing efficient adaptation without requiring full parameter updates.

\subsubsection{Supervised Fine-Tuning (SFT)}
\label{subsubsec:sft}

Supervised Fine-Tuning (SFT) further adapts the model by explicitly teaching it to map dermatological inputs to structured SOAP notes based on weakly supervised ground-truth examples. Unlike pre-training, which loosely guides the model, SFT provides complete input-output pairs, enabling the model to learn structured clinical reasoning patterns.

The SFT training objective minimizes the cross-entropy loss between the predicted SOAP note \(\hat{y}\) and the target SOAP note \(y\) given the multimodal input \(x\) (image and caption):

\begin{equation}
\mathcal{L}_{\text{SFT}} = - \sum_{i=1}^{n} y_i \log(\hat{y}_i)
\end{equation}

\noindent
where \(\mathcal{L}_{\text{SFT}}\) denotes the supervised fine-tuning loss, \(n\) is the number of tokens in the SOAP note, \(y_i\) is the ground-truth token at position \(i\), \(\hat{y}_i\) is the predicted probability for the \(i\)-th token, and \(x\) represents the multimodal input consisting of the lesion image and its corresponding clinical caption.

This loss function ensures that the generated notes are not only linguistically coherent but also structurally accurate and clinically reliable.

\subsection{Training Setup}
\label{subsec:training_setup}

The fine-tuning of the Vision-LLaMA 3.2 model is performed using Quantized Low-Rank Adaptation (QLoRA) with a low-rank dimension \( r = 8 \), a scaling factor \( \alpha = 16 \), and no dropout applied. LoRA modules are inserted into the query, key, value, output, gate, up, and down projections within the model's transformer blocks. Supervised Fine-Tuning (SFT) is conducted with a batch size of 8, using gradient accumulation over 4 steps and 10 warmup steps at the beginning of training. Fine-tuning is performed for 500 epochs with a linear learning rate scheduler, starting from an initial learning rate of \( 2 \times 10^{-4} \). The optimizer used is AdamW with 8-bit precision to improve memory efficiency. Pretraining to generate weakly supervised SOAP notes took approximately 9 hours, while the final fine-tuning process was completed in 1.5 hours on an NVIDIA A100 GPU with 80 GB of VRAM. Mixed-precision training with bfloat16 (bf16) format was employed to optimize memory utilization throughout the training process.

\subsection{Inference}
\label{subsec:inference}

At inference time, the fine-tuned Vision-LLaMA model receives a lesion image along with its corresponding clinical features, which were first converted into a clinical caption. The model then generates a structured SOAP note. Because the model has been fine-tuned on weakly supervised yet clinically reliable data, it generalizes effectively to new cases. This enables scalable and structured generation of dermatology SOAP notes, even in environments where expert annotations are limited or unavailable.

\section{Evaluation}


We evaluate the generated SOAP notes using both quantitative and qualitative methods. For quantitative evaluation we compare the generated notes against expert-annotated ground truth using standard and clinical-domain NLP metrics. Additionally, we introduce two clinical relevance metrics: \textbf{MedConceptEval} and \textbf{Clinical Coherence Score (CCS)}. The qualitative evaluation is based on the LLM-as-a-Judge approach \cite{zheng2023judging} with Flow-Judge-v0.1, where the model evaluates structure, readability, medical relevance, and consistency with conventional SOAP note standards.

\subsection{Quantitative Evaluation}



\subsubsection{MedConceptEval}

To evaluate clinical relevance, we introduce MedConceptEval, a semantic evaluation framework designed to assess the alignment of each SOAP note section with clinically validated concept sets. These concept sets, referred to as descriptor banks, are curated for six major dermatological classes derived from reputable clinical resources \cite{mayoclinic_melanoma_symptoms}. A language model is employed to extract relevant medical concepts and construct disease-specific keyword sets. Each section of the generated SOAP note is encoded using ClinicalBERT, and cosine similarity is computed against the corresponding descriptor bank. For each dermatological class, we calculate both average and maximum similarity scores per section across five representative cases. This method provides a robust, interpretable, and clinically grounded evaluation of the generated SOAP notes, ensuring alignment with disease-specific clinical terminology beyond surface-level keyword matching.


\begin{table}[htbp]
\centering
\scriptsize
\renewcommand{\arraystretch}{1.2}
\begin{tabular}{|p{2.5cm}|l|c|c|}
\hline
\textbf{Condition} & \textbf{Section} & \textbf{Avg Similarity} & \textbf{Max Similarity} \\
\hline
\multirow{4}{=}{Seborrheic Keratosis (SEK)}
& Subjective & 0.7768 & 0.8746 \\
& Objective  & 0.7952 & 0.8648 \\
& Assessment & 0.8168 & 0.8680 \\
& Plan       & 0.7764 & 0.8310 \\
\hline

\multirow{4}{=}{Nevus (NEV)}
& Subjective & 0.7786 & 0.8468 \\
& Objective  & 0.7786 & 0.8598 \\
& Assessment & 0.8006 & 0.8708 \\
& Plan       & 0.8626 & 0.8976 \\
\hline

\multirow{4}{=}{Melanoma (MEL)}
& Subjective & 0.7790 & 0.8624 \\
& Objective  & 0.7952 & 0.8676 \\
& Assessment & 0.8234 & 0.8770 \\
& Plan       & 0.8526 & 0.9036 \\
\hline

\multirow{4}{=}{Actinic Keratosis (ACK)}
& Subjective & 0.7354 & 0.8182 \\
& Objective  & 0.7844 & 0.8554 \\
& Assessment & 0.8092 & 0.8458 \\
& Plan       & 0.8400 & 0.8864 \\
\hline

\multirow{4}{=}{Squamous Cell Carcinoma (SCC)}
& Subjective & 0.7846 & 0.8360 \\
& Objective  & 0.7754 & 0.8360 \\
& Assessment & 0.7802 & 0.8398 \\
& Plan       & 0.7854 & 0.8596 \\
\hline

\multirow{4}{=}{Basal Cell Carcinoma (BCC)}
& Subjective & 0.7740 & 0.8464 \\
& Objective  & 0.7658 & 0.8182 \\
& Assessment & 0.7882 & 0.8220 \\
& Plan       & 0.7738 & 0.8212 \\
\hline
\end{tabular}
\caption{\textbf{MedConceptEval:} Semantic similarity between SOAP sections and curated clinical concept sets across six dermatologic conditions.}
\label{tab:medconcepteval}
\end{table}

In Table \ref{tab:medconcepteval}, the Assessment and Plan sections consistently achieved higher average similarity scores compared to Subjective and Objective sections, indicating stronger alignment with medically relevant concepts. Conditions like Melanoma and Nevus demonstrated particularly high alignment, with maximum similarity values exceeding 0.90 in the Plan section, suggesting that generated notes closely matched expert medical descriptors. However, slightly lower scores were observed for SCC and BCC, highlighting potential areas for improvement in capturing subtle clinical features for these cases.

\textbf{Statistical Significance Analysis:} We conducted a Two-Way ANOVA to evaluate the effects of SOAP note sections and dermatological conditions on the average MedConceptEval similarity scores. The results indicated a statistically significant main effect of the SOAP section \((F(3, 15) = 4.31, p = 0.022)\), demonstrating that semantic alignment varied across the Subjective, Objective, Assessment, and Plan sections. However, the effect of lesion type was not significant \((F(5, 15) = 1.44, p = 0.268)\), suggesting that the model’s semantic consistency was maintained across different skin conditions. Here, \(F(d_1, d_2)\) represents the F-statistic with \(d_1\) and \(d_2\) degrees of freedom for between-group and within-group variance, respectively. These results underscore the greater influence of SOAP note structure over disease type in shaping semantic alignment with clinical concepts.

\subsubsection{Clinical Coherence Score (CCS)}
We introduce the Clinical Coherence Score (CCS), a metric that evaluates the semantic alignment between the caption and the structured SOAP note sections. To compute this score, we utilize ClinicalBERT based contextual embeddings, which are specifically designed to capture clinical terminology and relationships. Unlike traditional lexical overlap metrics (e.g., ROUGE, BLEU), CCS provides a domain-specific signal of how faithfully the model retains and reflects the original clinical information in each section of the output. It does so by embedding both the caption and each SOAP section into a shared semantic space using ClinicalBERT and computing their cosine similarity.

We have compared the CCS for both our generated notes and dermatologists written notes as shown in table \ref{tab:ccs_comparison} which indicates that the LLM-generated SOAP notes exhibit consistently higher semantic alignment with the captions compared to the dermatologist provided handwritten notes. These scores suggests that our model is more skillful at capturing the terminology and phrasing present in the caption, likely due to its training in a shared embedding space with similar data. While, clinician written notes may introduce complex language or clinical reasoning not explicitly present in the caption, resulting in slightly lower CCS despite being more accurate.

This opens up an opportunity in research that reveals that LLM, still does not fully replicate the depth and complexity of hand-written notes. Bridging this gap could enable future models to match or even complement the quality, consistency, and clinical reasoning present in expert written documentation.

\textbf{Statistical Significance Analysis: }
To examine the impact of different note type and SOAP structure on semantic alignment, we performed a Two-Way ANOVA on CCS values across three lesion cases. The analysis revealed a statistically significant main effect of note type \((F(1, 19) = 57.53, p < 0.001)\), indicating that generated notes exhibited substantially higher semantic alignment with the caption than ground truth notes. In contrast, the effect of SOAP section was not significant \((F(3, 19) = 1.47, p = 0.254)\), suggesting that the variation in alignment did not differ meaningfully across the Subjective, Objective, Assessment, and Plan sections. These results shows the model generated notes are consistently coherent outputs across all sections of the SOAP note, surpassing human-written references in terms of semantic alignment with the input caption. Given that the reference notes were written by a domain expert, this finding suggests a promising avenue for future research, where LLM generated clinical documentation could eventually match or even complement the quality and consistency of expert-written notes.

\begin{table}[h]
\centering
\small
\resizebox{\linewidth}{!}{  
\begin{tabular}{|l|ccc|ccc|}
\hline
\textbf{SOAP} & \multicolumn{3}{c|}{\textbf{Generated notes}} & \multicolumn{3}{c|}{\textbf{Ground Truth notes}} \\
             & Case 1 & Case 2 & Case 3 & Case 1 & Case 2 & Case 3 \\
\hline
Subjective   & 0.9308 & 0.9306 & 0.9318 & 0.3335 & 0.7184 & 0.7475 \\
Objective    & 0.9168 & 0.8891 & 0.9334 & 0.5448 & 0.4235 & 0.7239 \\
Assessment   & 0.9178 & 0.8933 & 0.9208 & 0.4926 & 0.7266 & 0.7299 \\
Plan         & 0.8795 & 0.8842 & 0.8981 & 0.3899 & 0.5371 & 0.3368 \\
\hline
\textbf{{Average}} & \cellcolor{blue!15}{0.9112} & \cellcolor{blue!15}{0.8993} & \cellcolor{blue!15}{0.9210} & \cellcolor{red!15}{0.4402} & \cellcolor{red!15}{0.6014} & \cellcolor{red!15}{0.6501} \\
\hline
\end{tabular}
}
\caption{\textbf{Clinical Coherence Score:} Semantic alignment between the caption and each SOAP section for both generated and ground truth SOAP notes across three lesion cases.}
\label{tab:ccs_comparison}
\end{table}


Furthermore, we obtained expert-annotated SOAP notes for three different lesion images from a board-certified dermatologist. Each lesion image, along with its corresponding caption, was provided to the dermatologist to generate structured SOAP notes. These expert-written notes were treated as ground truth (reference) and are compared with our generated notes (candidate) to evaluate their alignment using a range of lexical and semantic metrics, including BLEU \cite{post2018call}, ROUGE \cite{lin2004rouge}, METEOR \cite{lewis2020retrieval}, CHRF++ \cite{popovic2017chrf++}, BERT Score \cite{zhang2019bertscore}, and ClinicalBERT Score \cite{huang2019clinicalbert}.

\begin{table*}[t]
\centering
\small
\resizebox{1.00\textwidth}{!}{%
\begin{tabular}{|l|cccc|cccc|cccc|}
\hline
\textbf{Metrics} & \multicolumn{4}{c|}{\textbf{Case 1}} & \multicolumn{4}{c|}{\textbf{Case 2}} & \multicolumn{4}{c|}{\textbf{Case 3}} \\
\cline{2-13}
& skin-SOAP & GPT-4o & Janus Pro & Claude & skin-SOAP & GPT-4o & Janus Pro & Claude & skin-SOAP & GPT-4o & Janus Pro & Claude \\
\hline
ROUGE-1       & \cellcolor{blue!15}{0.396} & 0.324 & 0.348 & 0.294 & 0.4183 & 0.399 & 0.397 & 0.460 & 0.3999 & 0.4360 & 0.351 & 0.480 \\
ROUGE-2        & 0.0827 & 0.0326 & 0.1228 & 0.0281 & \cellcolor{blue!15}{0.125} & 0.0966 & 0.1122 & 0.0671 & 0.0939 & 0.0909 & 0.1088 & 0.0805 \\
ROUGE-L     & 0.1748 & 0.1943 & 0.2347 & 0.1538 & \cellcolor{blue!15}{0.2614} & 0.2214 & 0.2299 & 0.2266 & 0.2181 & 0.2556 & 0.2229 & 0.2472 \\
METEOR  & \cellcolor{blue!15}{0.2221} & 0.1745 & 0.1804 & 0.1952 & \cellcolor{blue!15}{0.2495} & 0.1692 & 0.1728 & 0.2202 & 0.2276 & 0.2242 & 0.1968 & 0.2370 \\
CHRF++  & \cellcolor{blue!15}{44.91} & 44.90 & 37.25 & 38.92 & \cellcolor{blue!15}{43.78} & 42.39 & 41.64 & 42.30 & \cellcolor{blue!15}{47.515} & 45.71 & 41.75 & 47.33 \\
BERT (F1)  & \cellcolor{blue!15}{0.1223} & 0.0619 & 0.1144 & 0.0117 & 0.0974 & 0.1487 & 0.1460 & 0.1481 & 0.0770 & 0.2100 & 0.0994 & 0.2105 \\
Clinical BERT (F1) & \cellcolor{blue!15}{0.7750} & 0.7409 & 0.7348 & 0.6974 & 0.7609 & 0.7723 & 0.7550 & 0.7528 & 0.7890 & 0.8098 & 0.7573 & 0.7846 \\
\hline
\end{tabular}
}
\caption{Evaluation of generated SOAP notes across lexical (ROUGE, METEOR), character-level (CHRF++), and semantic F1-score (BERT, Clinical BERT) metrics for three cases. Models: \textbf{Our Approach (skin-SOAP)}, GPT-4o, Janus Pro Deepseek, Claude 3.7 Sonnet.}
\label{tab:all_model_eval_combined}
\end{table*}

\subsubsection{ROUGE}
ROUGE (Recall-Oriented Understudy for Gisting Evaluation) measures n-gram overlap between generated and reference notes. We report ROUGE-1 (unigrams), ROUGE-2 (bigrams), and ROUGE-L (longest common subsequence), capturing both lexical recall and structural consistency.


\subsubsection{METEOR}
METEOR (Metric for Evaluation of Translation with Explicit ORdering) incorporates synonymy and word stemming to better handle clinical paraphrasing. It is more tolerant to surface-level variations and is especially useful when evaluating partially reworded notes.

\subsubsection{CHRF++}
CHRF++ computes F-scores over character n-grams, offering robustness to minor lexical differences, such as pluralization, typos, or morphological variations (e.g., “carcinoma” vs. “carcinomas”). It is particularly well-suited for clinical domains where such variations are common.

\subsubsection{BERT Score}
BERT Score uses contextual embeddings from a pretrained language model (typically BERT-base) to compute semantic similarity between reference and generated notes. It evaluates whether generated tokens align semantically with reference tokens beyond exact matches.

\subsubsection{Clinical BERT Score}
To capture clinically grounded semantics, we use BERT with Clinical BERT, a model pretrained on large-scale clinical notes (e.g., MIMIC-III). This allows for more accurate evaluation of clinical relevance, capturing medical synonyms, abbreviations, and contextual language common in SOAP documentation.


In the Table \ref{tab:all_model_eval_combined}, skin-SOAP demonstrates consistently strong performance, especially on METEOR and CHRF++, indicating high fluency and surface-level coherence. While GPT-4o achieved slightly higher ROUGE-L in Cases 1 and 3, our method outperformed it in ROUGE-2 and METEOR, which better reflect phrasal and semantic overlap. Notably, our ClinicalBERT F1 scores are either the highest or on par with the best-performing model in each case, underscoring our model's superior alignment with clinical concepts. Here, the difference between the BERT and ClinicalBERT scores further highlights how clinically tuned models capture semantic relevance more effectively than general-purpose transformer models. Overall, our results are comparable with the other models and these results confirm the effectiveness of our weakly supervised multimodal approach for producing high-quality and clinically grounded SOAP notes.

\subsection{Qualitative Evaluation}
\subsubsection{LLM-as-a-Judge}

\begin{figure*}[t]
    \centering
    \includegraphics[width=0.90\textwidth]{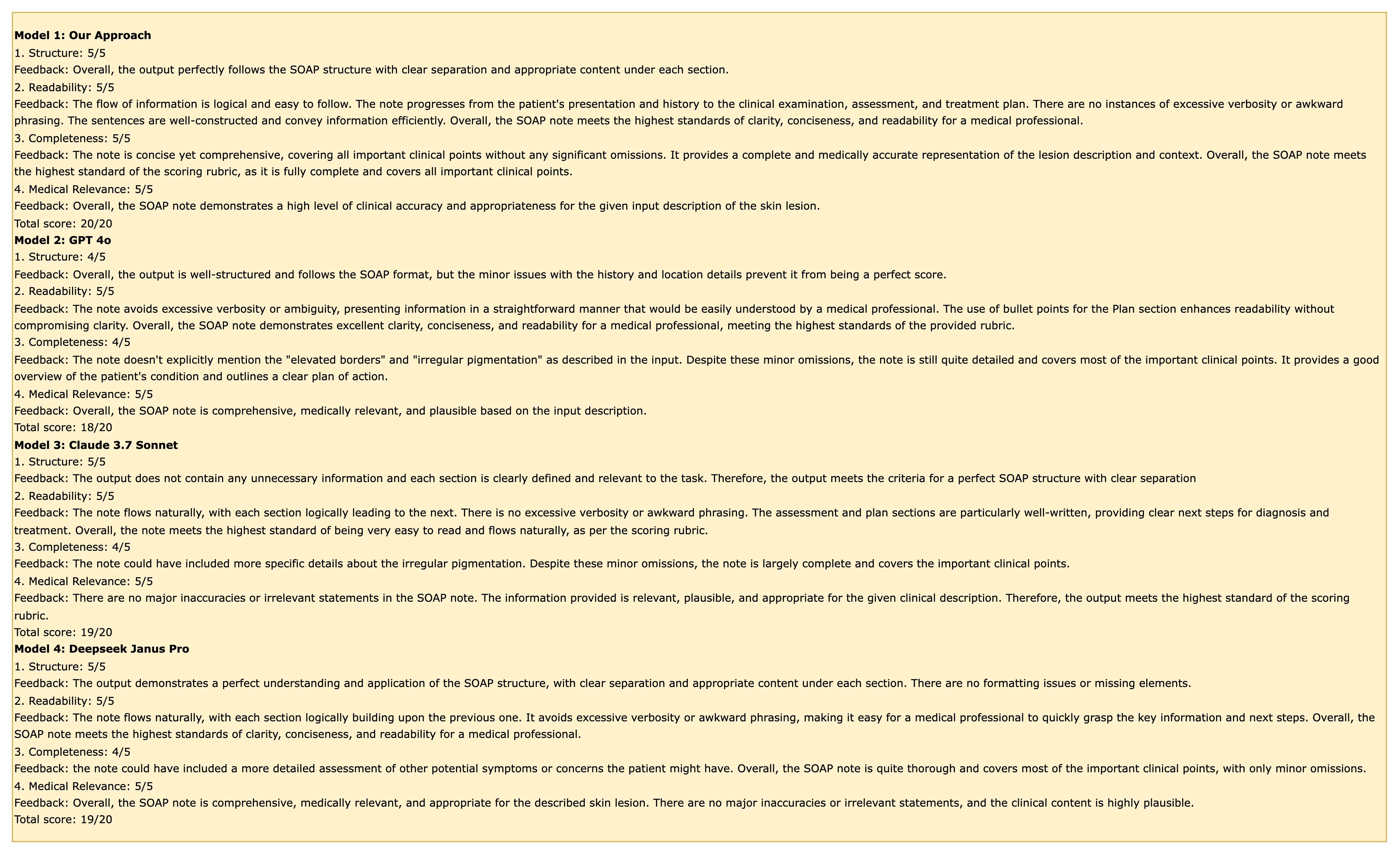}
    \caption{Comparison of Flow-Judge Feedback Across Four Language Models}
    \label{fig:LLM}
\end{figure*}


We conducted a qualitative evaluation using an LLM-as-Judge framework, as obtaining human verifiers is difficult, expensive. Specifically, we employed \textbf{Flow-Judge-v0.1} to assess each generated SOAP note across four criteria. According to the HuggingFace Judge Arena: Benchmarking LLMs as Evaluators, Flow-Judge (3.8B) is an open-source model developed by Flow AI, achieves the highest ELO rating of 1335, outperforming larger proprietary models such as GPT-4o (1320), Claude 3 Opus (1268), and Meta Llama 3.1 405B (1267). In our study, Flow-Judge rated each note on a 5-point Likert scale (1 = Poor, 5 = Excellent) based on the following evaluation criteria:

\begin{itemize} \item \textbf{Structure:} Does the clinical note correctly follow the structured SOAP format, with distinct and appropriate content under each section (\textbf{S}: Chief Complaint and Medical History; \textbf{O}: Examination findings and Observations; \textbf{A}: Investigations, Diagnosis, and Summary; \textbf{P}: Treatment Plan and Patient Education)? \item \textbf{Readability:} Is the language of the clinical note clear, concise, and readable for a medical professional without excessive complexity or ambiguity? \item \textbf{Completeness:} Does the clinical note cover all the key details described in the input lesion description and context, and address all aspects of the clinical scenario, ensuring that no critical details are overlooked? \item \textbf{Medical Relevance:} Is the clinical content of the SOAP note medically relevant, plausible, and appropriate given the input description of the skin lesion? \end{itemize}


This setup enabled a structured, blinded review of the generated outputs, allowing us to capture both technical and clinical quality dimensions beyond traditional quantitative scores. The total scores for each model were shown in Fig \ref{fig:LLM}, our approach achieved a perfect score of 20/20, with feedback highlighting its clear organization, clinical accuracy, and comprehensive coverage of all relevant details. In contrast, GPT-4o and the other baselines scored between 18/20 and 19/20. While their outputs were well-structured and readable, minor issues such as omissions of specific symptoms or less detailed assessments slightly impacted their completeness scores. Overall, Flow-Judge evaluations confirm that our framework delivers higher clinical fidelity and completeness than SOTA models, highlighting key strengths and improvement areas.

\section{Example of Generated SOAP Notes}

\begin{figure*}[t]
    \centering
    \includegraphics[width=0.7\textwidth]{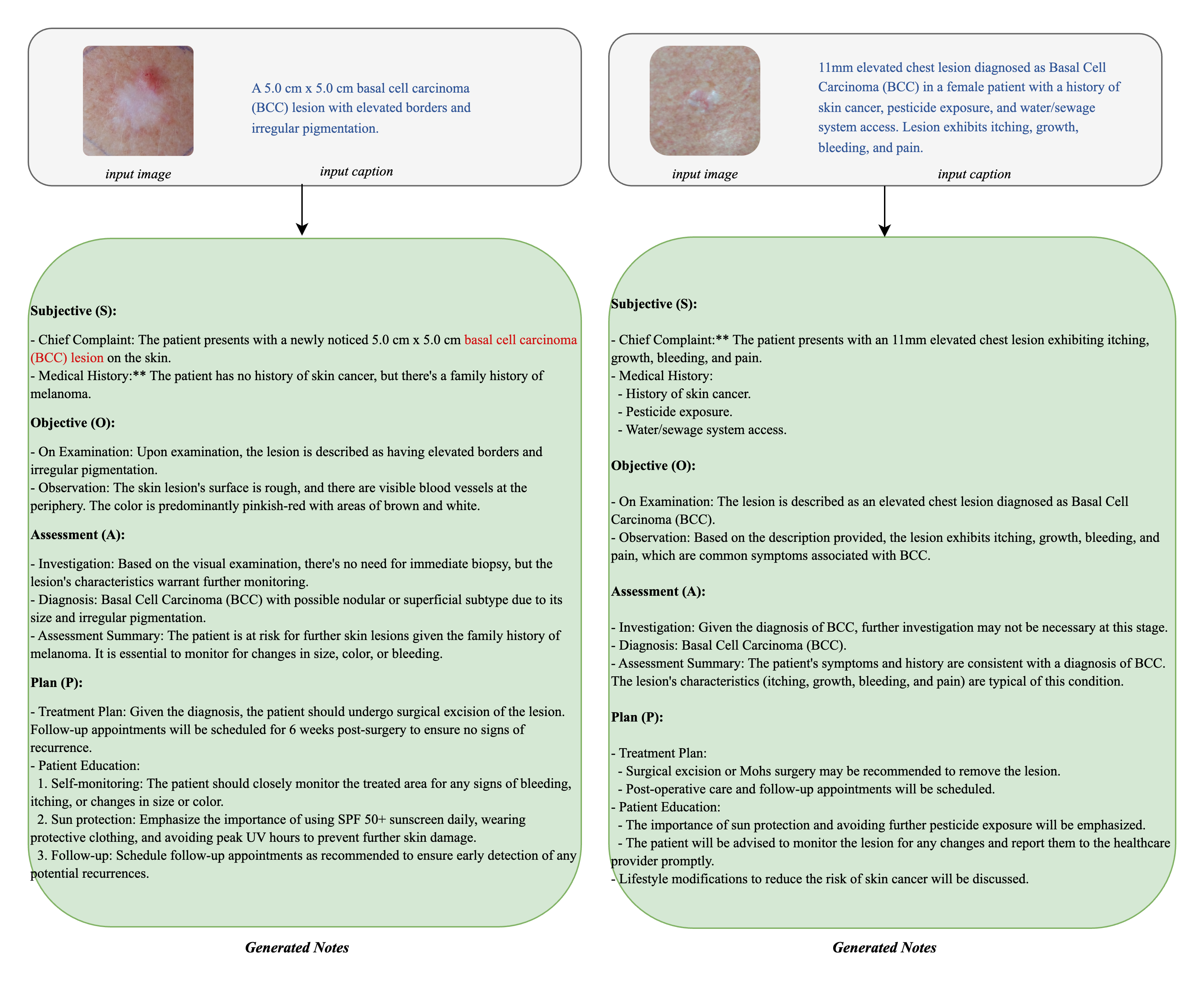}
    \caption{Examples (a) and (b) show structured SOAP notes generated by our proposed multimodal framework using an input lesion image and its corresponding caption.}
    \label{fig:result}
\end{figure*}


Figure~\ref{fig:result} shows two representative SOAP notes generated by our model using a lesion image and its caption. These examples demonstrate the framework’s ability to produce clinically structured and coherent documentation aligned with the SOAP format. In both cases, the model integrates visual and textual inputs to generate comprehensive notes.

Figure~\ref{fig:result}(a) illustrates a structural error, where the diagnosis of BCC is incorrectly placed in the Chief Complaint section instead of the Assessment section. In contrast, Figure~\ref{fig:result}(b) shows a correctly formatted note, with symptoms captured under Chief Complaint and diagnosis placed appropriately in Assessment. These examples highlight the model’s potential to learn documentation structure, while also revealing areas for improvement.

\section{Conclusion}
\label{sec:conc}

In this work, we presented, skin-SOAP, a weakly supervised multimodal framework for generating clinically structured SOAP notes from limited dermatologic inputs. By leveraging generative language model generated captions, retrieval-augmented knowledge integration, and fine-tuning a Vision-LLaMA model with weak supervision, this study reduces dependence on large-scale expert annotations while maintaining strong clinical relevance and structural coherence. Through both qualitative and quantitative evaluations, including our proposed metrics such as MedConceptEval and Clinical Coherence Score (CCS), we demonstrated that our method produces high-quality, clinically meaningful notes, advancing the development of scalable and reliable clinical documentation systems. Ultimately, our framework has the potential to accelerate dermatology clinical workflows, reduce time-to treatment, and improve overall patient care.

\section{Limitations and Future Work}
\label{sec:limit}

While our proposed framework shows strong promise for structured SOAP notes generation, it has few limitations. The quality of the generated notes remains dependent on the accuracy of the retrieved domain-specific knowledge, which may introduce biases or propagate incomplete information. Additionally, we utilized only a single dataset, as variations in metadata across different sources posed challenges for standardization. Our evaluation was further constrained by a small set of expert-annotated samples, restricting large-scale validation. Like most generative models, our approach may hallucinate when faced with ambiguous or unfamiliar inputs. Although retrieval-augmented generation helps reduce this risk by grounding outputs in clinical knowledge, additional precautions are needed for real-world deployment. In Future work, we will focus on expanding to more diverse datasets, and incorporating human-in-the-loop refinement strategies. Furthermore, developing evaluation benchmarks that capture the progression of clinical reasoning across multiple encounters and support decision making could significantly improve the utility of automated SOAP note generation in real-world healthcare settings.

\bibliographystyle{named}
\bibliography{ijcai25}
\end{document}